\documentclass[lettersize,journal]{IEEEtran}
\usepackage{amsmath,amsfonts}
\usepackage{algorithmic}
\usepackage{algorithm}
\usepackage{array}
\usepackage[caption=false,font=normalsize,labelfont=sf,textfont=sf]{subfig}
\usepackage{textcomp}
\usepackage{stfloats}
\usepackage{url}
\usepackage{verbatim}
\usepackage{graphicx}
\usepackage{cite}
\usepackage[dvipsnames]{xcolor}  
\usepackage{multirow}
\usepackage{listings}  
\usepackage{float}
\usepackage[citecolor=blue,linkcolor=blue]{hyperref}

\begin{document}
	
\lstset{
	language = Python,
	basicstyle = \footnotesize\ttfamily,           
	breaklines = true,                  
	numberstyle = \small,               
	keywordstyle = \color{blue},            
	commentstyle =\color{black}\large\textbf,        
	stringstyle = \color{red!100},          
	frame = single,                  
	showspaces = false,                 
	columns = fixed,                    
	morekeywords = {*,...,.,=},                
	deletendkeywords = {compile}
}

\title{4K-HAZE: A Dehazing Benchmark with 4K Resolution Hazy and Haze-Free Images}

\author{Zhuoran Zheng and Xiuyi Jia, ~\IEEEmembership{Member,~IEEE}}




\maketitle

\begin{abstract}
Currently, mobile and IoT devices are in dire need of a series of methods to enhance 4K images with limited resource expenditure.
The absence of large-scale 4K benchmark datasets hampers progress in this area, especially for dehazing. 
The challenges in building ultra-high-definition (UHD) dehazing datasets are the absence of estimation methods for UHD depth maps, high-quality 4K depth estimation datasets, and migration strategies for UHD haze images from synthetic to real domains.
%
To address these problems, we develop a novel synthetic method to simulate 4K hazy images (including nighttime and daytime scenes) from clear images, which first estimates the scene depth, simulates the light rays and object reflectance, then migrates the synthetic images to real domains by using a GAN, and finally yields the hazy effects on 4K resolution images.
We wrap these synthesized images into a benchmark called the 4K-HAZE dataset.
%
%
Specifically, we design the CS-Mixer (an MLP-based model that integrates \textbf{C}hannel domain and \textbf{S}patial domain) to estimate the depth map of 4K clear images, the GU-Net to migrate a 4K synthetic image to the real hazy domain.
The most appealing aspect of our approach (depth estimation and domain migration) is the capability to run a 4K image on a single GPU with 24G RAM in real-time (33fps).
Additionally, this work presents an objective assessment of several state-of-the-art single-image dehazing methods that are evaluated using the 4K-HAZE dataset.
At the end of the paper, we discuss the limitations of the 4K-HAZE dataset and its social implications.
The code URL at \url{https://github.com/zzr-idam/4KDehazing}.
\textcolor{red}{Since all images are in UHD resolution (\textbf{arxiv} has a limit on the size of manuscript that can be uploaded), this manuscript is only a primer; the complete manuscript at \url{https://www.researchgate.net/publication/368426400_4K-HAZE_A_Dehazing_Benchmark_with_4K_Resolution_Hazy_and_Haze-Free_Images}.}
\end{abstract}


\section{Introduction}

\label{sec:intro}
%
With the large-scale application of \textbf{U}ltra-\textbf{H}igh-\textbf{D}efinition (UHD) images on handheld devices, haze degrades the visual quality of UHD images the outdoor.
%
Most single-image dehazing methods address the visual degradation of low-resolution images, but these methods encounter the obstacle on running a 4K image.

%
%
The obstacle in the UHD image dehazing field is that the dehazing models are trained on low-resolution images, and these models lack the knowledge (finer details of the structure, denser pixels in local areas) to make inferences about UHD image processing.
To address the problem, we built a UHD dehazing dataset containing both daytime and nighttime scenes, named 4K-HAZE.
%
%
Specifically, we propose the CS-Mixer model to estimate the depth of the 4K resolution image to assist the atmospheric scattering model to generate a non-homogeneous 4K haze image (haze distribution has a non-homogeneous character in many real scenes), then we use GU-Net to render a scene closer to the real haze based on a GAN scheme.
Note that CS-Mixer is trained on a customized 4K depth estimation dataset ( including 500 pairs of images).
The 4K depth map uses guide filtering and Adobe's \textit{Depth Scanner} (\url{https://pan.baidu.com/s/11kwFtYPLP-QsUBOgJj3KGQ?pwd=1234}).
This study introduces 4K-HAZE which represents the first 4K image dehazing dataset with non-homogeneous hazy and haze-free (ground-truth) paired images.
This work has two very noteworthy aspects: \textbf{a)} the blurring effect of the image comes from the estimation of the blurring kernel of the real-world image; \textbf{b)} both CS-Mixer and GU-Net can accurately conduct depth estimation and image degradation on a UHD resolution.
%
%
\begin{figure}[t] 
	\begin{center}
		\begin{tabular}{@{}c@{}}
			\includegraphics[width = 0.48\textwidth]{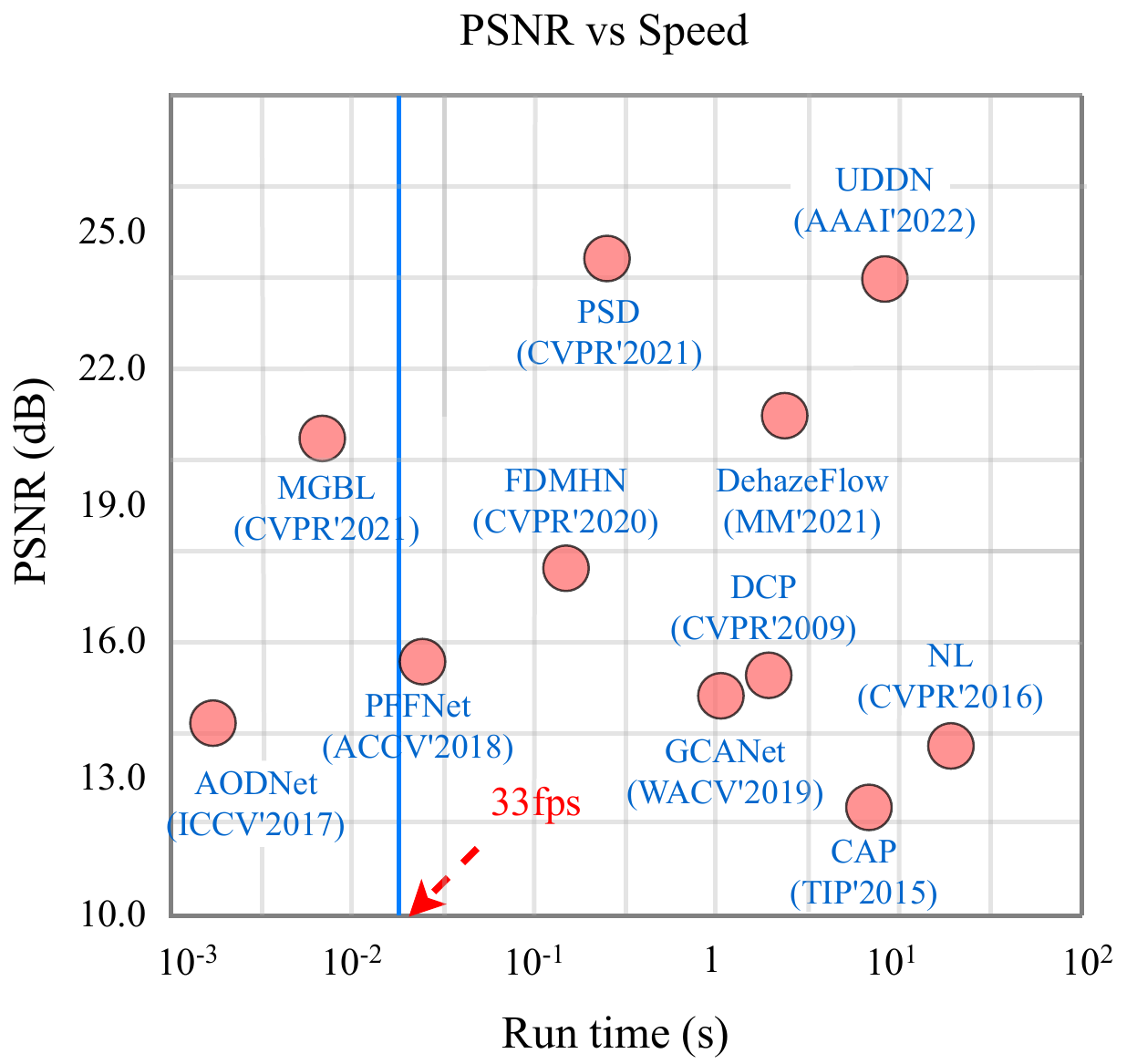}                   
		\end{tabular}
	\end{center}
	\vspace{-4mm}
	\caption{Speed and accuracy trade-off of the state-of-the-art dehazing methods on the 4K-HAZE dataset. \textbf{\textcolor{blue}{The blue line}} denotes the real-time approach for 4K resolution images at 33 fps.
	These methods can directly train full-resolution images on an NVIDIA Tian RTX3090 GPU shader with 24G RAM.
		}
	\vspace{-6mm}
	\label{ms-psnr}
\end{figure}
4K-HAZE contains over 30,000 pairs of 4K images recorded outdoors.
Additionally, this work presents a comprehensive evaluation of several state-of-the-art single-image dehazing methods~\cite{mgbl, psd, dcp, nl, cap, aod, fdmhn, pff, tang2019single, li2021dehazeflow, liu2022image}, that were objectively evaluated on our new dataset.
As shown in Figure~\ref{ms-psnr}, the analyzed single-image dehazing techniques have been assessed quantitatively using two traditional metrics: PSNR and run time.
DCP~\cite{dcp}, CAP~\cite{cap}, and NL~\cite{nl} have achieved grand success in image dehazing, but their performances are currently far inferior to deep learning methods.
AODNet~\cite{aod}, FDMHN~\cite{fdmhn}, PFFNet~\cite{pff}, GCANet~\cite{tang2019single},  MGBL~\cite{mgbl}, DehazeFlow~\cite{li2021dehazeflow}, UDDN~\cite{liu2022image} and PSD~\cite{psd} can be dehazed in real time on a single 24G RAM RTX3090 GPU.
Among them, MGBL~\cite{mgbl} and PSD~\cite{psd} achieved outstanding performance in terms of speed and accuracy.
Note that all of the above methods act on the full-resolution images ($3 \times 3840 \times 2160$) during the training stage (these methods do not conduct resize, random cropping, mixture so on).
To summarize, the main contributions of this paper are summarized as follows:
\begin{itemize}
\item We design a synthetic 4K image with a haze pipeline, which can handle a 4K image by using CS-Mixer and GU-Net in an end-to-end manner.
\item We propose CS-Mixer and GU-Net, both of which have the core component of capturing global perception capability by rolling the spatial and channel domains; to the best of our knowledge this is the first strategy to remove the CNN stem and feed it directly into the MLPs.
\item Our built 4K-HAZE provides well-rounded knowledge (including the nighttime and the daytime scenes) of existing dehazing methods, in addition to significant improvements in downstream tasks.
\end{itemize}
\begin{figure*}[t] 
	\begin{center}
		\begin{tabular}{@{}c@{}}
			\includegraphics[width = 1\textwidth]{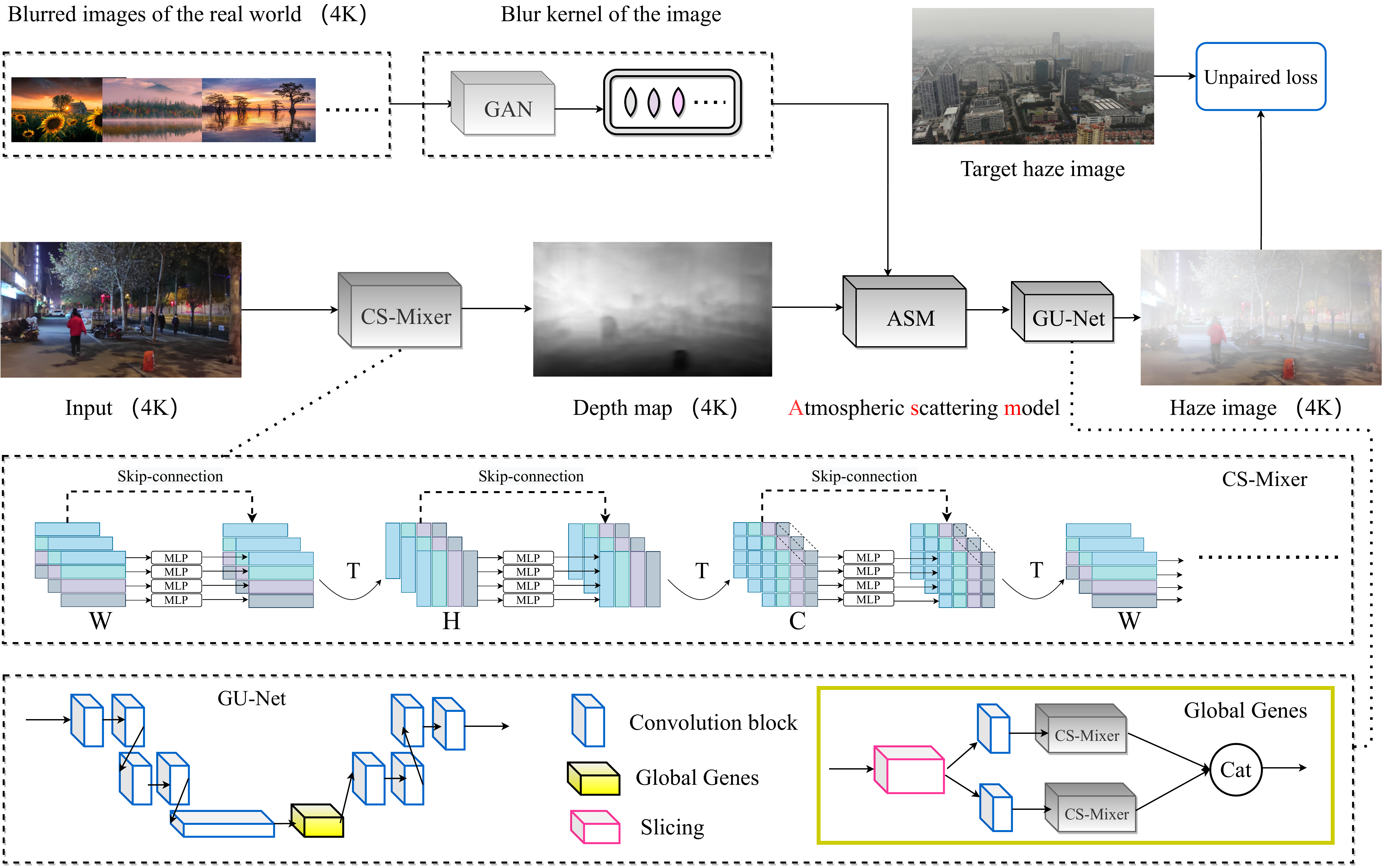}                   
		\end{tabular}
	\end{center}
	\caption{The architecture and configuration of the proposed single 4K hazy-image synthesis method, which consists of four parts. First, we develop the CS-Mixer to estimate a 4K depth map. Then we adopt the Dehazing optical model (Atmospheric scattering model) to generate a 4K non-uniform hazy image.
		Next, we extract blur kernels from real-world 4K blurred images with the help of a trained extractor and apply the extracted blur kernels randomly on the 4K non-uniform haze image. Finally, the 4K non-uniform haze image migrates from the synthetic domain to the domain of the real hazed image with the help of GU-Net and the unpaired loss function. Note that our algorithm downsamples the images ($3 \times 256 \times 256$) at the entrance of CS-Mixer and GU-Net, respectively, followed by upsampling back to the raw resolution at the exit of the model. Specifically, the image is output as a full--esolution feature map though CS-Mixer or GU-Net. The feature maps are then multiplied with the input full-resolution image ($3 \times 3840 \times 2160$).}
	\vspace{-2mm}
	\label{framework}
\end{figure*}

\section{Limitations and Social Impact}
%
%
Since it is currently very expensive to create high-quality UHD depth estimation datasets, we use Adobe's built-in algorithm to estimate depth maps of arbitrary resolution as ground truth.
For the depth map being estimated, we conduct an enhancement in HD mode, such as using a guided filtering algorithm to sharpen the edge information of the depth map.
Although this strategy can construct a high-quality 4K image with haze to some extent, the depth estimation method is still inadequate on devices with limited computing power.
In the future, we will provide dehazing datasets for Under Display Cameras environment and dusty weather.
The data set for the above conditions will likewise collect a basket of daytime environments versus nighttime environments.
Notably, we record images that do not involve sensitive personal or confidential information.
%
\section{Conclusions}
\label{sec:con}
In this paper, we build a novel dataset for UHD image dehazing. 
Dehazing models can be trained in UHD full-resolution images without cropping.
The key to our method is the use of a CS-Mixer and a GU-Net that can help obtain a high-quality 4K depth map and a 4K hazy image.
Quantitative and qualitative results show that PSD performs favorably against state-of-the-art dehazing methods, and can generate visually-pleasing results on real-world hazy images.

\clearpage

\bibliographystyle{ieee_fullname}
\bibliography{e}

\end{document}